\documentclass[conference]{IEEEtran}
\IEEEoverridecommandlockouts
\usepackage{cite}
\usepackage{amsmath,amssymb,amsfonts}
\usepackage{algorithmic}
\usepackage{graphicx}
\usepackage{textcomp}
\usepackage{xcolor}
\def\BibTeX{{\rm B\kern-.05em{\sc i\kern-.025em b}\kern-.08em
    T\kern-.1667em\lower.7ex\hbox{E}\kern-.125emX}}

\usepackage{times}
\usepackage{soul}
\usepackage{url}
\usepackage[hidelinks]{hyperref}
\usepackage[utf8]{inputenc}
\usepackage[small]{caption}
\usepackage{booktabs}
\usepackage{algorithm}
\usepackage[switch]{lineno}
\usepackage{pgfplots}
\usepackage{subcaption}
\usepackage{array}

\definecolor{color1}{HTML}{D81B60}
\definecolor{color1dark}{HTML}{9E1345}
\definecolor{color2}{HTML}{1E88E5}
\definecolor{color3}{HTML}{FFC107}
\definecolor{color4}{HTML}{004D40}
\definecolor{color5}{HTML}{FE6100}

\newcommand{\greenai}{\textcolor[rgb]{0,0.5,0}{Green~AI}}

\newcommand{\researchq}[1]{$RQ_{#1}$}
\newcommand{\materialurl}{https://anonymous.4open.science/r/inference-batching-BE1F}

\newcommand{\energywtscattercombiV}[2]{
\begin{tabular}{@{}c@{}}
\begin{tikzpicture}
    \fontsize{12pt}{14.4pt}\selectfont 12pt
    \begin{axis}[
        width  = 8cm,
        height = 7.7cm,
        major x tick style = transparent,
        ybar,
        bar width=14pt,
        ymin=25,
        ymax=60,
        xmin=0,
        xmax=5,
        ymajorgrids = true,
        ylabel = {Max Wait Time (s)},
        ylabel style = {yshift=-6pt},
        xlabel = {Energy / Image  (J)},
        xlabel style = {yshift=-4pt},
        scaled y ticks = false,
        legend style={at={(0.02,0.98)},anchor=north west},
    ]
    \addplot[
        scatter/classes={128={color4}},
        scatter, mark=*, only marks, 
        scatter src=explicit symbolic,
        nodes near coords*={\Label},
        every node near coord/.append style={xshift=2pt,yshift=0pt,anchor=west},
        visualization depends on={value \thisrow{label} \as \Label}
    ] table [meta=class]{#2};
        \legend{f=128}
    \end{axis}
\end{tikzpicture}
\\
\begin{tikzpicture}
    \fontsize{12pt}{14.4pt}\selectfont 12pt
    \begin{axis}[
        width  = 8cm,
        height = 7.7cm,
        major x tick style = transparent,
        ybar,
        ymin=0,
        ymax=12,
        xmin=0,
        xmax=5,
        bar width=14pt,
        ymajorgrids = true,
        ylabel = {Max Wait Time (s)},
        ylabel style = {yshift=-10pt},
        xlabel = {Energy / Image  (J)},
        xlabel style = {yshift=-4pt},
        scaled y ticks = false,
        legend style={at={(0.02,0.98)},anchor=north west,column sep=1ex},
        legend columns=2
    ]
    \addplot[
        scatter/classes={16={color1},32={color2},64={color3}},
        scatter, mark=*, only marks, 
        scatter src=explicit symbolic,
        nodes near coords*={\Label},
        every node near coord/.append style={xshift=2pt,yshift=0pt,anchor=west},
        visualization depends on={value \thisrow{label} \as \Label}
    ] table [meta=class]{#1};
        \legend{f=16,f=32,f=64}
    \end{axis}
\end{tikzpicture}
\end{tabular}
}

\begin{document}

\title{Batching for \textcolor[rgb]{0,0.5,0}{Green~AI} - An Exploratory Study on Inference}

\author{
\IEEEauthorblockN{Tim Yarally\IEEEauthorrefmark{1}, Lu\'is Cruz\IEEEauthorrefmark{1},
Daniel Feitosa\IEEEauthorrefmark{2},
June Sallou\IEEEauthorrefmark{1}, Arie van Deursen\IEEEauthorrefmark{1}}
\IEEEauthorblockA{\IEEEauthorrefmark{1}Delft University of Technology, The Netherlands - {timyarally@hotmail.com,  \{ l.cruz, j.sallou, arie.vandeursen \}@tudelft.nl}}
\IEEEauthorblockA{\IEEEauthorrefmark{2}University of Groningen, The Netherlands - {d.feitosa@rug.nl}}
}

\maketitle

\begin{abstract}
The batch size is an essential parameter to tune during the development of new neural networks. Amongst other quality indicators, it has a large degree of influence on the model's accuracy, generalisability, training times and parallelisability. This fact is generally known and commonly studied. However, during the application phase of a deep learning model, when the model is utilised by an end-user for inference, we find that there is a disregard for the potential benefits of introducing a batch size.
In this study, we examine the effect of input batching on the energy consumption and response times of five fully-trained neural networks for computer vision that were considered state-of-the-art at the time of their publication.
The results suggest that batching has a significant effect on both of these metrics. Furthermore, we present a timeline of the energy efficiency and accuracy of neural networks over the past decade. We find that in general, energy consumption rises at a much steeper pace than accuracy and question the necessity of this evolution. Additionally, we highlight one particular network, \textit{ShuffleNetV2}~(2018), that achieved a competitive performance for its time while maintaining a much lower energy consumption. Nevertheless, we highlight that the results are model dependent.
\end{abstract}


\section{Introduction}
\label{sec:introduction}

Sustainability has emerged as a challenging optimisation problem in the AI research community. The community is building the most powerful models with a colossal number of parameters, but their massive energy footprint is an issue yet to be solved. For example, the state-of-the-art GPT-3 model has 175 billion parameters and has been estimated to require more than 1 Gigawatt-hour of energy to be trained~\cite{patterson2022carbon}. While the numerous applications enabled by these models are impressively innovative, there is a growing concern about the sustainability of taking these models to production.  

A new field, dubbed \greenai, is rising to address this concern~\cite{SLRGreenAI}. The initial contributions in \greenai{} consist of positional papers that are calling for a new research agenda~\cite{bender2021dangers,strubell2019energy,schwartz2020green}. This involves the measurement and reporting of energy consumption next to accuracy, but also the appreciation of research efforts that do not necessarily rely on enterprise-sized data or training budgets. Nonetheless, with enlarging models and more complex training, the energy demand grows considerably. Microsoft's partnership with OpenAI to build a dedicated supercomputer with 10,000 GPUs~\cite{langston2020MSHPC} is a recent example. Thus, we advocate for an urgent need for boosting \greenai{} to support the ever-growing AI energy demands.

From a study by Facebook AI, we learn that at least 50\% of the operational carbon cost of machine learning tasks can be attributed to inference~\cite{wu2022sustainable}. Fully trained models can be deployed to a huge number of independent devices that collectively process a lot of data.
Moreover, devices that act as hosts to the neural networks do not necessarily have the same computational power as the machine used for training. In the case of mobile devices, battery life also becomes a factor. For these use cases, efficiency is essential. To explore this problem, some studies specifically focus on developing computation-efficient models~\cite{zhang2018shufflenet,ma2018shufflenet}. Other works focus on developing strategies for multiple-model selection, based on the idea a diverse set of models can meet different energy and performance requirements~\cite{ijcai2020p473}. Based on the same principle, by creating multiple instances of cascading models with increasing complexity, energy-intensive models can be called only when deemed necessary~\cite{ijcai2018p302}.

We argue that one should not only optimise for training~\cite{yarally2023uncovering} and development, but consider the complete life-cycle of a neural network. The batch size (i.e., the number of input data samples that are processed at one time during inference) is one of the most important hyperparameters to tune during the training phase. It has implications on the model accuracy and generalisability~\cite{kandel2020effect}, training times and parallelisability~\cite{smith2017don}, etc. During inference, however, there is no dataset available that can be divided into batches. Instead, the incoming stream of requests depends on some external factor that provides input. Hence, any attempt to process data in batches of a specific size inadvertently introduces a form of delay to the response. This is an important difference from the training phase because the GPU always exerts some amount of power even when idle. 

In this study, we analyse this two-way optimisation problem between the energy consumption and the response time during inference. In addition, we present a timeline of state-of-the-art neural networks and compare them in terms of their energy consumption. We chose to focus on computer vision networks because of their wide range of solutions and diverse approaches as the field has evolved. 
In summary, we strive to answer the following research questions:

\begin{enumerate}
    \item[\researchq{1}:] How does batch inference affect the energy consumption of computer vision tasks under different frequencies of incoming requests?
    \item[\researchq{2}:] How has the energy efficiency of computer vision models evolved in the last decade?
\end{enumerate}

The methods and tools used in this study are accounted for in Section~\ref{sec:methods}. In Section~\ref{sec:doe}, we go over the experiment that was devised to collect the delay and the energy consumption for different experimental settings. This section also explains the data collection for the neural network energy timeline. The results of the experiment are presented in Section~\ref{sec:results}, and we analyse and elaborate on these in Section~\ref{sec:discussion}. Finally, we take note of the threats to our study in Section~\ref{sec:t2v} and wrap up our findings and recommendations with the conclusion in Section~\ref{sec:conclusion}.

\section{Research Methods}
\label{sec:methods}
The goal of this study is to examine the effect of
input batching on the energy consumption and response times of five fully-trained neural networks.

\subsection{Case Selection}
\label{subsec:case}
We first select networks that are considered \textit{state-of-the-art} (SotA) at their publication period. We assess this criterion based on the accuracy reported in the original publication and SotA leaderboards such as from ``Papers with Code\footnote{\url{https://paperswithcode.com/sota/image-classification-on-imagenet}}''. Next, we collect their pre-trained models provided by Pytorch\footnote{\url{https://pytorch.org/vision/stable/models.html}}:
\begin{itemize}
    \item \textbf{AlexNet} (2014)~\cite{krizhevsky2012imagenet}
    \item \textbf{DenseNet} (2016)~\cite{huang2017densely}
    \item \textbf{ShuffleNetV2} (2018)~\cite{ma2018shufflenet}
    \item \textbf{VisionTransformer} (2020)~\cite{dosovitskiy2020image}
    \item \textbf{ConvNext} (2022)~\cite{liu2022convnet}
\end{itemize}

The reason we choose these five networks in particular is because their initial publication dates are spread out evenly in the past decade. This not only provides a good variety of different network designs, but it also facilitates \researchq{2}, where we attempt to compare the energy consumption of modern neural networks to their predecessors. It should also be noted that these models are designed for image classification. We choose this problem space because image recognition is a canonical deep learning challenge. 

All experiments are performed on a single GeForce GTX-1080 GPU\footnote{\url{https://www.nvidia.com/nl-nl/geforce/10-series/}}. The stop condition for any run mentioned in this study is a fixed amount of processed image classification requests. Because inference is reliant on external providers for incoming requests, the time it takes to receive a certain amount of requests can vary a lot. To make a fair comparison of the differences in energy consumption, we assume regular streams of incoming requests in this study.

\subsection{Experimental Tooling}
We develop a testbed in Python to automate the data collection. The testbed is publicly available in an open source repository to enable reproducibility\footnote{\url{\materialurl}}. The software provides a simulated queue that creates image classification tasks at a frequency that can be configured manually. Requests are then pulled from the queue and collected in a batch with configurable size. These batches are fed to the neural networks. Apart from the five networks mentioned in Section~\ref{subsec:case}, our tool is immediately compatible with any image vision model that Pytorch provides or any custom model that is built using the same framework. We highly encourage experimenting with different architectures and reporting the results.

\subsection{Data Collection}
For this study, we are interested in two quality metrics: the average energy usage per image classification and the maximum response time, meaning the time between a user submitting a task and receiving an answer. As mentioned before, we assume that the stream of incoming requests is about constant. This entails that the time between any two requests will be roughly the same for a fixed frequency. 

We obtain the power usage of the GPU by querying the NVIDIA System Management Interface\footnote{\url{https://developer.nvidia.com/nvidia-system-management-interface}} every 10 milliseconds. The total energy consumption can then be computed as a factor of time and the average power. Finally, this amount is divided by the total number of images to calculate the desired metric. For this study, we do not factor out the idle consumption of the GPU because idling is an important part of the experiment. By increasing the batch size, we inherently increase idle times as well. The experiment is meant to show whether this increase in batch size and response time improves the energy efficiency or not.

The maximum response time is determined by providing each incoming classification task with a timestamp. This timestamp is resolved as soon as the request is handled, and the program keeps track of the longest time in memory.

\section{Experiments}
\label{sec:doe}
In the following section, we describe the setup of the experiment in detail. We use the results of this single experiment to answer both research questions (\researchq{1} \& \researchq{2}). 

\subsection{Batching During Inference}
During the image vision training phase, a neural network processes thousands upon thousands of images. To parallelise this task and employ more of the available GPU power, these images are often processed in batches. During inference, however, when we look at image classification in a practical setting, the usual dataset is replaced by an irregular stream of incoming requests. We refer to the number of images that come in per second as the \textit{frequency}. If we choose to perform inference in larger batch sizes, depending on the frequency, we might have to wait for a batch to fill up before passing it on to the network and this increases the response time to the user. In a nutshell, this is the game that we attempt to optimise: the trade-off between energy consumption and wait time.

\begin{figure}
    \centering
    \includegraphics[width=0.95\linewidth]{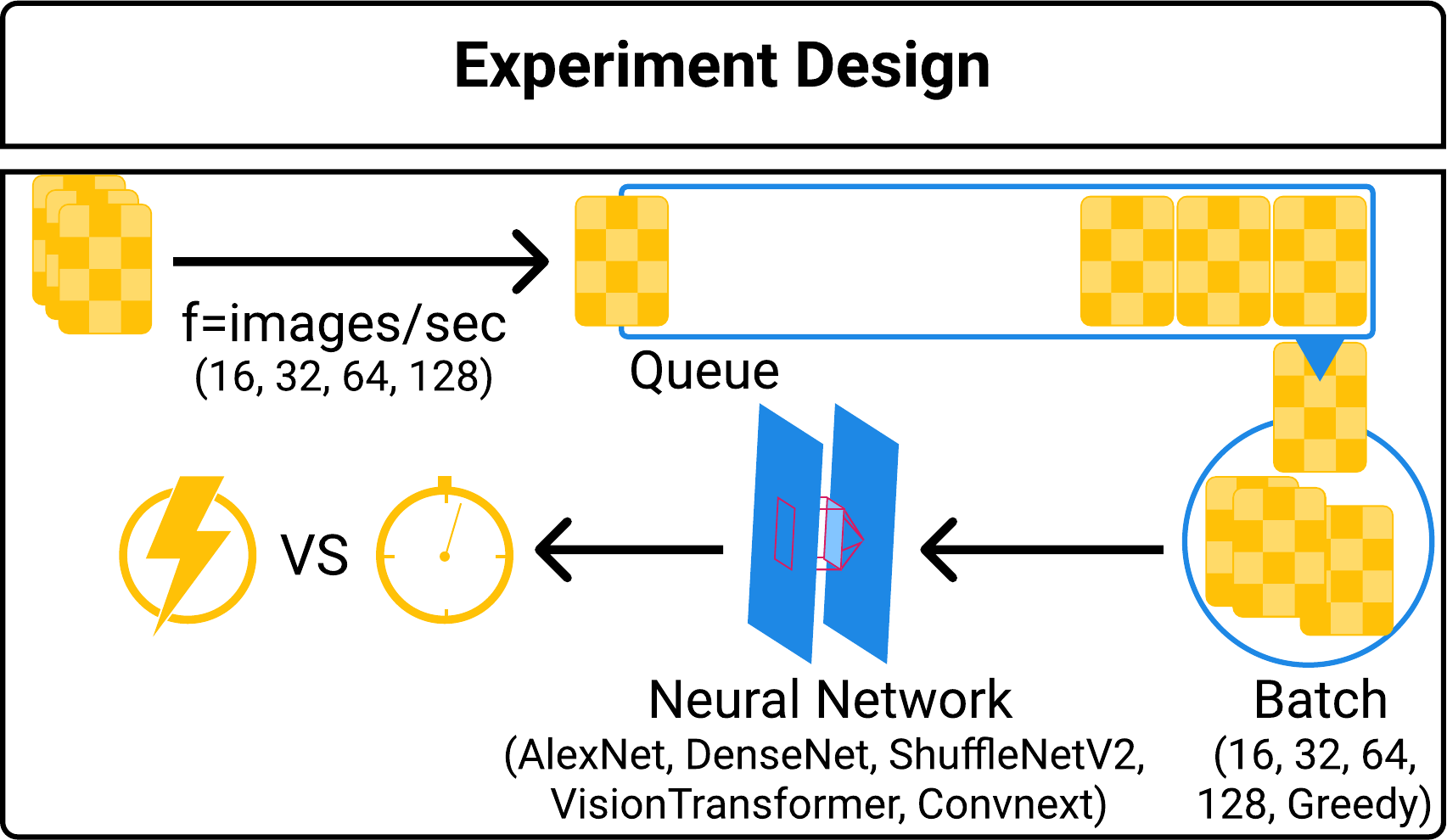}
    \caption{Inference experiment diagram}
    \label{fig:inf_exp}
\end{figure}

To carry out this experiment, we simulate a queue that receives incoming image classification requests. These images are then passed to a neural network using some batching strategy. The setup is as follows: we compare four different frequencies (16, 32, 64 \& 128); the five different networks mentioned in Section~\ref{sec:methods} and five batching strategies (16, 32, 64, 128 \& Greedy). This amounts to 100 different experimental configurations, where each configuration is represented by a triplet: \texttt{<frequency, network~model, batching~strategy>}. The greedy batching strategy is the baseline to which the other batch sizes are compared. Greedy in our simulation means all the images in the queue are passed to the network as soon as it becomes available\footnote{The maximum greedy batch size is set to 128 to avoid out-of-memory issues}. The flowchart in Figure~\ref{fig:inf_exp} displays this entire process in a graphical format.

Each configuration triplet  comprises one \textit{run}, which continues until $2^{13}$ image classification tasks have been requested and processed. To minimise variance with regards to process times, we cycle through three arbitrary images of roughly the same size. Note that because we use the request count as the termination criterion, we cannot compare settings with different frequencies to each other in terms of energy consumption. This is because as long as the GPU can keep up with the incoming stream, the frequency will determine for how long the simulation will continue and during idle periods, the GPU will still exert power. Therefore, we expect that the absolute energy consumption of low-frequency simulations will be greater than that of high-frequency ones. For this reason, we compare only the results that were accumulated using the same frequency with each other when formulating our answers to \researchq{1}.

\subsection{Image Vision Energy Timeline}
To answer the second research question (\researchq{2}), we calculate the average energy consumption per image over all five batching strategies for each combination of neural network and simulation frequency. This amounts to four values per model or 20 data points in total. We present these results in a bar chart in Section~\ref{sec:results}.

\section{Results}
\label{sec:results}
In this section, we present the results from the inference batching experiment. 

\subsection{Batching During Inference}
\begin{figure*}
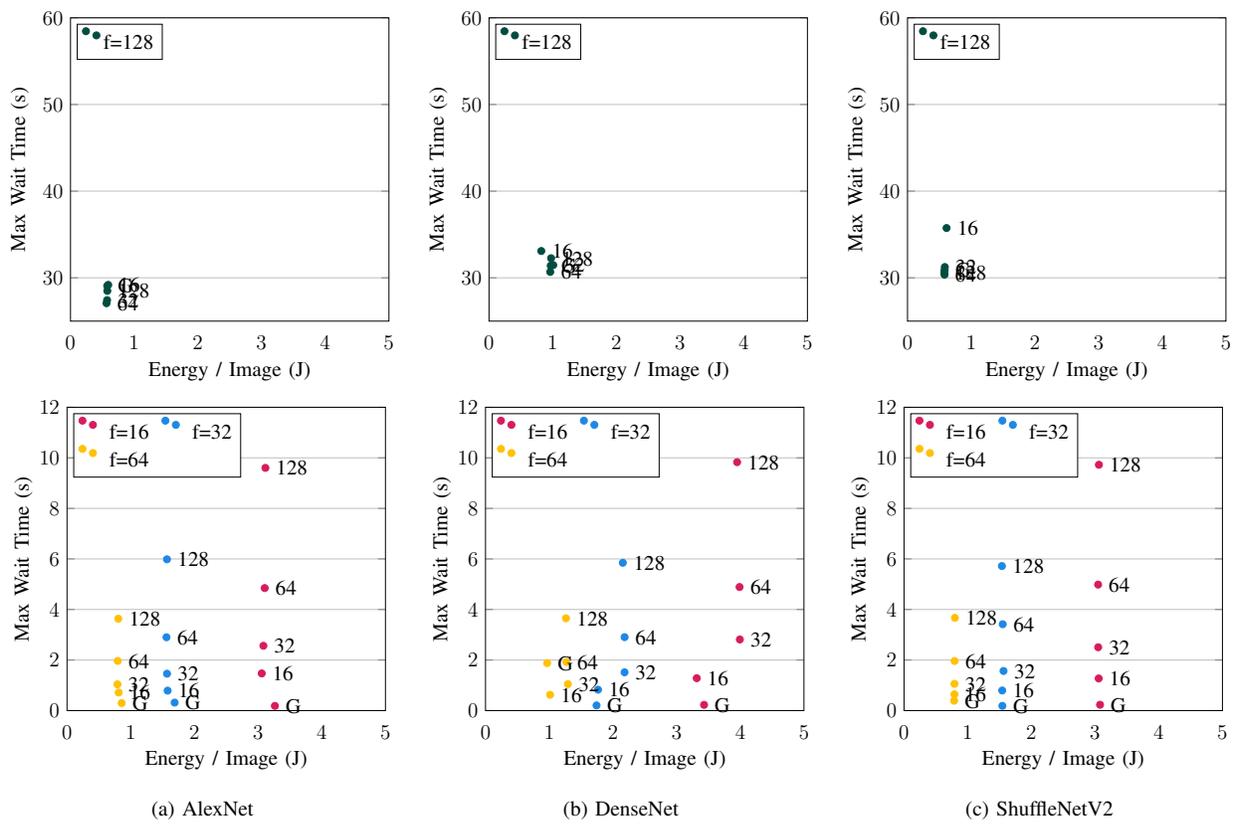
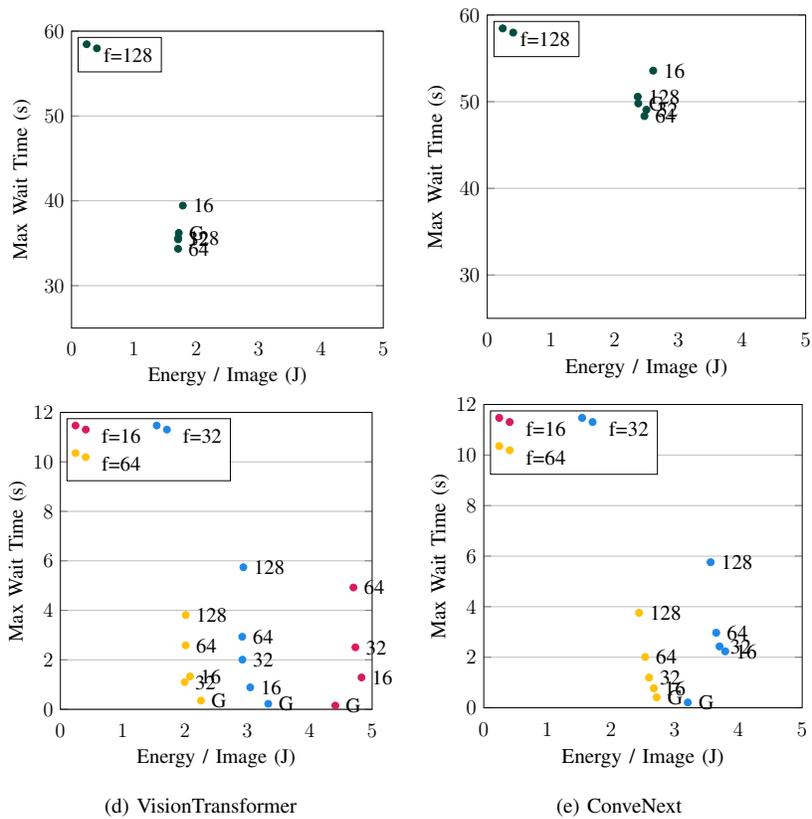

    \centering
    \begin{subfigure}[b]{0.3\textwidth}
        \centering
        \resizebox{\linewidth}{!}{%
        \energywtscattercombiV{data/energy-wt-alexnet-min.dat}{data/energy-wt-alexnet.dat}%
        }
        \caption{AlexNet}
        \label{fig:ew-alexnetL}
    \end{subfigure}
    \begin{subfigure}[b]{0.3\textwidth}
        \centering
        \resizebox{\linewidth}{!}{%
        \energywtscattercombiV{data/energy-wt-densenet-min.dat}{data/energy-wt-densenet.dat}%
        }
        \caption{DenseNet}
        \label{fig:ew-densenetL}
    \end{subfigure}
    \begin{subfigure}[b]{0.3\textwidth}
        \centering
        \resizebox{\linewidth}{!}{%
        \energywtscattercombiV{data/energy-wt-shufflenet-min.dat}{data/energy-wt-shufflenet.dat}%
        }
        \caption{ShuffleNetV2}
        \label{fig:ew-shufflenetL}
    \end{subfigure}
    \\[3ex]
    \begin{subfigure}[b]{0.3\textwidth}
        \centering
        \resizebox{\linewidth}{!}{%
        \energywtscattercombiV{data/energy-wt-vit-min.dat}{data/energy-wt-vit.dat}%
        }
        \caption{VisionTransformer}
        \label{fig:ew-vitL}
    \end{subfigure}
    \begin{subfigure}[b]{0.3\textwidth}
        \centering
        \resizebox{\linewidth}{!}{%
        \energywtscattercombiV{data/energy-wt-convnext-min.dat}{data/energy-wt-convnext.dat}%
        }
        \caption{ConveNext}
        \label{fig:ew-convnextL}
    \end{subfigure}
    \caption{Energy consumption per unit inference vs the maximum response time. For each sub-figure, the upper plot show low frequency simulations (16, 32, 64) and lower plot show high frequency simulations (128).}
    \label{fig:ew-tradeof}
\end{figure*}

For each of the five image vision models (i.e. AlexNet, DenseNet, ShuffleNetV2, VisionTransformer \& ConvNext), we perform a trade-off analysis concerning the average energy consumption per processed image and the maximum wait time in the queue. The results are visualised in the scatter plots from Figure~\ref{fig:ew-tradeof}. In these plots, the x-axis represents the average energy consumption in Joules for processing a single image. The y-axis shows the maximum time from when a user submits an image until she receives a response. Furthermore, there are four different classes that each corresponds to a different setting of the simulation. For a class \textit{f=X}, \textit{X} represents the frequency of the incoming image requests, e.g. the class \textit{f=32} will have 32 image requests every second. Finally, the labels next to each data point correspond to the size of the batches (i.e., 16, 32, 64, 128 and \textit{G}), where \textit{G} refers to the greedy batching strategy (i.e., all the images in the queue are sent as soon as it becomes available).

\begin{table}
    \centering
    \resizebox{\linewidth}{!}{%
    \begin{tabular}{lrrr}
        \toprule
        Model               &   Batch size 1-2 (W)  &   Batch size 128 (W)  &    Difference (\%)\\ 
        \midrule
        AlexNet             &   $\pm$65.0           &   87.3                &    $\pm$34.3 \\
        DenseNet            &   72.5                &   163.1               &    124.9 \\
        ShuffleNetV2        &   $\pm$65.0           &   87.1                &    $\pm$33.9 \\
        VisionTransformer   &   76.2                &   185.8               &    144.0 \\
        ConvNext            &   93.1                &   166.4               &    78.6 \\
        \bottomrule
    \end{tabular}%
    }
    \caption{GPU peak power in Watts (W) differences for small and large batch sizes}
    \label{tab:peak-power}
\end{table}

There are several things that we can observe from these scatter plots. First of all, every network responds to batching differently. AlexNet (Figure~\ref{fig:ew-alexnetL}) and ConvNext (Figure~\ref{fig:ew-convnextL}) both clearly benefit from batching as the greedy strategy is almost always the least energy efficient. Two exceptions are the frequency 32 simulation for ConvNext and the frequency 128 simulation in general. The latter is easy to explain if we consider the average batch size of the greedy strategy. For frequencies 16, 32 and 64, this ranges from 1 to 7 images per batch, whereas for the 128 frequency simulation, the average batch size lies around 122. Given the positive effect of larger batch sizes, we can understand why the greedy strategy would perform better in high-frequency scenarios.

The VisionTransformer (Figure~\ref{fig:ew-vitL}) also generally runs more efficiently for larger batch sizes. For this model, the exception can be found in the frequency 16 simulation. Here we find that the greedy strategy, with an average batch size of 1.0004, is the most energy-efficient.

For another interesting observation we direct our attention to the scatterplot for ShuffleNetV2 in Figure~\ref{fig:ew-shufflenetL}. We find that there is virtually no horizontal spread in the points, which suggests that the model's efficiency does not depend on the batch size. This belief is enforced if we also consider the average peak power of the GPU while processing a batch of images. For all the other models, there is a large difference in peak power for processing a small batch of images versus a large one. This does not hold for ShuffleNetV2, which can be seen in Table~\ref{tab:peak-power}. This table shows that not only ShuffleNetV2, but also AlexNet have a relatively small change in peak power for different batch sizes.

Finally we look at the results for DenseNet in Figure~\ref{fig:ew-densenetL}. Out of all five networks, the observed behaviour for DenseNet is the most contradictory. We find that it performs the most efficiently for smaller batch sizes regardless of the simulation frequency. Because the greedy strategy takes very small batch sizes (1-2) for frequencies 16-64, we find that greedy is actually a very energy-efficient strategy for DenseNet.

Nevertheless, we cannot make any design decisions based on these results without considering the second metric in the scatter plots. Although the greedy strategy is often the least energy-efficient, we see that it consistently achieves the lowest wait times. For the frequencies 16 through 64 simulations, the graphs show that the greedy strategy results in near-instant response times while increasing the batch size introduces a maximum delay between 1 and 15 seconds. For the high-frequency simulation, we observe a shift in this trend. Since the GPU is not quite able to process all the images as soon as they enter the queue, a bottleneck is formed. This results in higher wait times in general and we find that the smallest batch size of 16 is the least favourable in this case. Across all models, the batch size of 64 is the most optimal with regard to the maximum wait time. 

\subsection{Image Vision Energy Timeline}
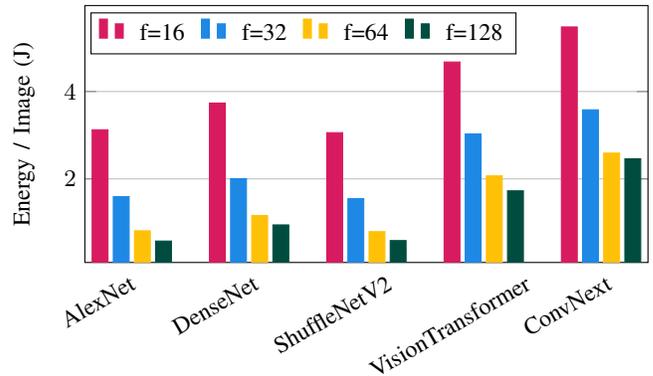
\begin{figure}
    \centering
    \begin{tikzpicture}
    \fontsize{9pt}{10.8pt}\selectfont 9pt
    \begin{axis}[
        width  = 0.5*\textwidth,
        height = 5cm,
        major x tick style = transparent,
        ybar,
        bar width=6pt,
        ymajorgrids = true,
        scaled y ticks = false,
        ylabel = {Energy / Image  (J)},
        ylabel style = {yshift=-12pt},
        symbolic x coords={AlexNet,DenseNet-121,ShuffleNetV2,ViT-b-16,ConvNext-base},
        xtick = data,
        xticklabels={AlexNet,DenseNet,ShuffleNetV2,VisionTransformer,ConvNext},
        xticklabel style={rotate=30,yshift=10pt,xshift=-9pt},
        legend style={at={(0.01,0.81)},anchor=south west,column sep=1ex},
        legend columns=-1
    ]
        \addplot[style={color1,fill=color1,mark=none}] table [x=model, y=f16, col sep=tab]{data/network_comparison.dat};
        \addplot[style={color2,fill=color2,mark=none}] table [x=model, y=f32, col sep=tab]{data/network_comparison.dat};
        \addplot[style={color3,fill=color3,mark=none}] table [x=model, y=f64, col sep=tab]{data/network_comparison.dat};
        \addplot[style={color4,fill=color4,mark=none}] table [x=model, y=f128, col sep=tab]{data/network_comparison.dat};
        \legend{f=16,f=32,f=64,f=128}
    \end{axis}
\end{tikzpicture}
     \caption{Energy comparison of different image vision models throughout the years}
    \label{fig:energy-comp}
\end{figure}

For the second part of this experiment, we take a step back to compare the overall energy consumption of the five models to each other. Figure~\ref{fig:energy-comp} shows the average energy required to process a single image in four different simulations. This average comes from the summation of the energy consumption for all the batch sizes for one such simulation. The models on the x-axis are in a specific order, which is not necessarily an increasing one in terms of energy efficiency. The models on the left and their respective papers were published before the models on the right. This creates an intuition for how energy efficiency evolves over time. The chart shows that there is a positive linear relationship between energy consumption and publication date. The exception to this trend is ShuffleNetV2, which, in terms of energy efficiency, is on the same level as AlexNet.

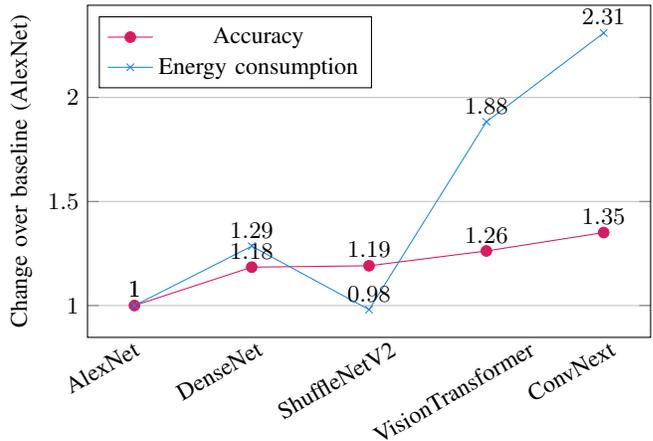
\begin{figure}[h]
    \centering
    \begin{tikzpicture}
        \fontsize{9pt}{10.8pt}\selectfont 9pt
        \begin{axis}[
            width  = 0.5*\textwidth,
            height = 6cm,
            major x tick style = transparent,
            ymajorgrids = true,
            ylabel = {Change over baseline (AlexNet)},
            ylabel style = {yshift=-10pt},
            symbolic x coords={AlexNet,DenseNet-121,ShuffleNetV2,ViT-b-16,ConvNext-base},
            xtick = data,
            xticklabels={AlexNet,DenseNet,ShuffleNetV2,VisionTransformer,ConvNext},
            xticklabel style={rotate=30,yshift=9pt,xshift=-9pt},
            scaled y ticks = false,
            legend style={at={(0.02,0.75)},anchor=south west}
        ]
            \addplot[mark=*, , nodes near coords, nodes near coords style={color=black}, style={color1}] table [x=model, y=acc, col sep=tab]{data/energy-accuracy.dat};
            \addplot[mark=x, nodes near coords, nodes near coords style={color=black}, style={color2}] table [x=model, y=energy, col sep=tab]{data/energy-accuracy.dat};
            \legend{Accuracy,Energy consumption}
    
        \end{axis}
    \end{tikzpicture}
    \caption{Relative changes in accuracy and energy consumption compared to AlexNet}
    \label{fig:norm-changes}
\end{figure}

It would not be fair to look at this graph without considering the improvements in accuracy that the newer models achieve. In Figure~\ref{fig:norm-changes}, we highlight the relative changes in accuracy and energy consumption from every network compared to AlexNet, which was published first. The energy consumption is based on the results from this study and the accuracy refers to the achieved top 1 accuracy on the ImageNet dataset\footnote{\url{https://paperswithcode.com/sota/image-classification-on-imagenet}}. From this graph we can conclude that since 2012, the energy consumption has seen a steep increase of 131\% and this trend does not start to fall off. Accuracy, on the other hand, has improved by 35\%. Also notice that despite ShuffleNetV2's energy efficiency, it does not seem to sacrifice anything in terms of performance.

\section{Discussion}
\label{sec:discussion}
In this section, we reflect and elaborate on the results as presented in Section~\ref{sec:results}. First, we consider the trade-off between energy and wait time to formulate an answer to \researchq{1}. After that, to answer \researchq{2}, we examine the collected energy consumption of the five networks. To further explain our findings, we look at the inner mechanisms and design principles of the five image vision models.

\subsection{Batching During Inference}
What we learn from the results might be somewhat unexpected. We did not find one recommended batch size or even an indication that reduces energy consumption in all cases. Instead, we find that each network behaves differently under varying batch sizes. Nonetheless, for some of the networks, the potential gain in energy efficiency cannot be ignored. 

AlexNet was published in 2012 and at that time it revolutionised the field of image vision. The architecture of the model is simple, with only five convolutional layers~\cite{krizhevsky2012imagenet}. ConvNext is a more modern CNN that incorporates design choices from classical CNNs like AlexNet and ResNet, that have been presented in the past decade~\cite{liu2022convnet}. Even the base model is quite a bit more complex than AlexNet, containing four different blocks for a total of 36 convolutional layers\footnote{\url{https://pytorch.org/vision/main/\_modules/torchvision/models/convnext.html}}. Nonetheless, both of these models are pure CNNs that do not rely on any special tricks. If we compare the results now, we find that there is a bias towards larger batch sizes as opposed to the small batch sizes of the greedy strategy in the low-frequency simulations. For the high-frequency simulation, where the inference becomes the bottleneck, the greedy strategy will process larger batch sizes, decreasing the energy consumed per image. Because we do not expect large data centres to experience this bottleneck, generally speaking, we can conclude that purely convolution-based models will benefit from performing inference in fixed batches ($>$16) rather than using a greedy strategy. For AlexNet, a batch size of 32 seems ideal because it limits the maximum wait time and the difference in energy consumption with the larger batch sizes is very small. For ConvNext, the largest batch size is nearly always the most energy-efficient, therefore we recommend a batch size of 128 (or even larger) when solely considering energy consumption. The increase in wait time should be evaluated per use case. 

The VisionTransformer is the only network that does not rely on convolution. Nevertheless, we find that batching almost always results in lower energy consumption when compared to the greedy strategy. Again we recommend a batch size of 32 for the same reason as with AlexNet. The difference in energy consumption with the largest batch sizes is small, so here we can afford to optimise for the wait time. Because we do not evaluate any other transformers in this study, it cannot be guaranteed that these results will generalise. However, since the VisionTransformer was designed to closely resemble the architecture of a ``standard transformer'' as used in NLP~\cite{dosovitskiy2020image}, we can make an educated guess that this will be the case.

ShuffleNetV2 and DenseNet are the anomalies in this experimentation. For ShuffleNet, we find that the energy consumption is completely invariant from the size of the batches. The wait times still scale as expected, therefore greedy batching is the most optimal inference strategy for this network. The energy consumption of DenseNet does differ per batching strategy, but this time with a bias towards smaller batch sizes. This network seems to consume the least amount of energy with batches of size $\leq$16. This means that for the low-frequency simulations, the greedy strategy is optimal. To get an intuition as to why this may be the case, we look at the architecture of DenseNet. In regular CNNs, the output of one layer is passed on only to the next layer. In DenseNet, all the layers are \textit{densely connected}, which means that any layer receives the output from all the preceding layers~\cite{huang2017densely}. All the layers remain occupied until an image has been completely processed by the network. One can imagine that this translates poorly to the parallel processing of multiple images.

Now that we have established how each network responds differently to batching and why that makes it difficult to provide recommendations, we move to answer to \researchq{1}: \textit{``How does batch inference affect the energy consumption for image vision tasks under different frequencies of incoming requests?''}

We find that in some cases, batching of the requests has a positive effect on the energy consumption of a neural network. However, there are strong exceptions to this observation. Our recommendation to AI practitioners is therefore as follows: When preparing newly trained networks for practical application, one should consider the batch size as an optimisation parameter that needs to be tuned. First, we establish whether the network runs more efficiently on small or larger batch sizes and then we tweak the batch size to lower values until the system adheres to the tolerated response time for the use case. 

\subsection{Image Vision Energy Timeline}
In terms of accuracy and energy consumption, the timeline that we have presented in Figure~\ref{fig:energy-comp} looks consistent and predictable. It also presents a critical problem: Although the innovations between 2012 and the present have led to impressive advancements in our neural networks and their precision, the potential gains in this regard are starting to diminish. From that, we formulate our answer for \researchq{2}: \textit{``How has the energy efficiency of image vision models evolved in the last decade?''}

Modern image vision models consume more than twice as much energy as earlier iterations and although these models demonstrate better performance, the gains in accuracy are limited. It is not surprising that we find this to be the case. Accuracy (or a similar measure) is the metric that currently defines what is ``state-of-the-art''~\cite{schwartz2020green}, in fact, many challenges and benchmarks only request a submission of the top-1 or top-5 accuracy. Some leaderboards do focus on cost or energy consumption\footnote{\url{https://dawn.cs.stanford.edu/benchmark/}}, but these are far and few between. If more challenges would accept submissions of new models where energy consumption is considered as a primary objective alongside accuracy, we can create opportunities for \greenai{} research. The proof that competitive models can also be efficient is already there. We established before that the ShuffleNetV2 architecture manages to break the increasing energy trend without bowing down to its predecessors in terms of accuracy. We look at the 2018 publication of ShuffleNetV2 to find out how this was accomplished~\cite{ma2018shufflenet}. The authors mention that most neural network design is guided by an indirect metric of the computational complexity: the number of floating-point operations (FLOPs). However, FLOPs only account for a part of the equation. The direct metric, speed, is also influenced by other processes like memory access. ShuffleNetV2 was designed with this mindset, to optimise for the direct metric of computational complexity rather than an indirect one. This goal of designing a fast network coincidentally resulted in a network that is also energy-efficient. In the same paper, the authors present a collection of four guidelines for efficient network design:
(1) Equal channel width minimises memory access cost (MAC);
(2) Excessive group convolution increases MAC;
(3) Network fragmentation reduces degree of parallelism; and
(4) Element-wise operations are non-negligible.

Many of these guidelines focus on the reduction of memory access during classification tasks. This could be an interesting starting point for future research in \greenai{}.

\section{Threats to Validity}
\label{sec:t2v}
In this section, we go through potential threats to the internal, external and construct validity, as well as the reliability.  

\textbf{Internal validity} regards the extent to which evidence supports cause-effect claims.
During early experimentation, we noticed that the GPU was idling on a higher power output for the first few minutes. Because this influenced the average energy consumption for some of the configurations, we introduced a warm-up phase. Before starting a new simulation and logging the energy consumption, we allowed the GPU to ``warm up'' by passing 256 batches of 32 images through the respective model. This factors out most of the inconsistencies. 

\textbf{External validity} addresses the extent to which our results can be generalised to broader contexts.
We mentioned before that the results collected in this study do not grant opportunities for firm recommendations and guidelines. Because we found a strong deviation in how different neural networks are influenced by batch inference, we can hardly claim that our findings will generalise well to other types of models. As such, our main contribution is not on the empirical results, but on the finding that a correct batching strategy will improve the overall energy efficiency and should therefore be tuned accordingly.

\textbf{Construct validity} concerns how well our indicators represent the intended object of study.
The main factor that hurts the construct validity is how accurately our simulation mirrors a real scenario. For the experiments, we assumed a constant workload with little to no deviation. In practice, one would expect a more erratic stream of incoming requests, with some periods of complete downtime. Naturally, it is undesirable to hold an unfilled batch while nothing new is coming in, so there should be some maximum time since the last request to avoid that. Nevertheless, our focus was not on optimising this simulation, but on investigating the energy efficiency of different batching strategies. Even in a more realistic scenario, the deviations in energy consumption that we observed should remain the same.

\textbf{Reliability} regards the extent to which the study can be replicated with the same observed results.
A single developer worked on accumulating the results presented in this study, but all the involved authors reviewed and approved the entire process. The complete reproduction package is available online\footnote{\url{\materialurl}}. This repository contains the source code that can be run to reproduce the results for any of the models from Section~\ref{sec:doe} or a different one provided by the Pytorch library\footnote{\url{https://pytorch.org/vision/stable/models.html}}.

\section{Conclusion}
\label{sec:conclusion}
In this study, we examined the energy efficiency of different neural networks that have been presented in the past decade. We simulated how these networks could be employed in a practical setting and extracted the optimal batching strategies for each. We learned that there is no one size fits all solution for recommending a batching strategy (\researchq{1}). 

AlexNet and ConvNext both operate more efficiently when using fixed batch sizes as opposed to greedy batching. Our results suggest a batch size of 32 for AlexNet and 128 (or larger) for ConvNext. Because of their classical architecture, we expect these results to generalise well to other pure CNN-based models. 

For the VisionTransformer, we find a similar result. A batch size of 32 appears to be the sweet spot in terms of GPU utilisation. A smaller batch size hurts the energy efficiency and a larger one does not provide any improvements. For future research, it would be interesting to repeat this experiment and evaluate more transformer-based models to see if these results generalise well.

The graphs from ShuffleNetV2 show little to no deviation in the energy consumption for different batching strategies. Based on these results we draw the conclusion that this neural network is batch size invariant with regard to the energy consumption. As such, the greedy strategy is the most optimal because it limits the maximum response time.

Finally, the results for DenseNet highlight why we chose to evaluate each network separately. Larger batch sizes actively hurt the energy efficiency of this model, therefore the greedy strategy is the most optimal one.

Furthermore, we presented an energy efficiency timeline in Figure~\ref{fig:energy-comp}. In general, we find that the energy consumption of modern neural networks has increased steadily in the last ten years (\researchq{2}). ConvNext, the most recent publication, consumes more than twice as much energy as the revolutionary AlexNet from 2012. Nevertheless, our timeline has an irregularity that holds a great opportunity. ShuffleNetV2 is the only model in our timeline that does not adhere to the increasing energy trend. Additionally, when compared to its predecessors AlexNet and DenseNet, we find that ShuffleNetV2 does not perform any worse. We looked at the design principles that were considered when developing this network and argue that future work should incorporate the views and guidelines presented in the corresponding publication. 

\bibliographystyle{splncs04}
\bibliography{refs}

\end{document}